# DUAL-STRATEGY IMPROVEMENT OF YOLOV11N FOR MULTI-SCALE OBJECT DETECTION IN REMOTE SENSING IMAGES

Zhu Shuaiyu[1], Sergey Ablameyko[1,2]

[1]*Belarusian State University, Minsk, 220030, Republic of Belarus*

[2]*United Institute of Informatics Problems, National Academy of Sciences of Belarus, Minsk, 220012 Republic of Belarus.*

Abstract—Satellite remote sensing images pose significant challenges for object detection due to their high resolution, complex scenes, and large variations in target scales. To address the insufficient detection accuracy of the YOLOv11n model in remote sensing imagery, this paper proposes two improvement strategies. Method 1: (a) a Large Separable Kernel Attention (LSKA) mechanism is introduced into the backbone network to enhance feature extraction for small objects; (b) a Gold-YOLO structure is incorporated into the neck network to achieve multi-scale feature fusion, thereby improving the detection performance of objects at different scales. Method 2: (a) the Gold-YOLO structure is also integrated into the neck network; (b) a MultiSEAMHead detection head is combined to further strengthen the representation and detection capability for small and multi-scale objects.

To verify the effectiveness of the proposed improvements, experiments are conducted on the DOTAv1 dataset. The results show that, while maintaining the lightweight advantage of the model, the proposed methods improve detection accuracy (mAP@0.5) by 1.3% and 1.8%, respectively, compared with the baseline YOLOv11n, demonstrating the effectiveness and practical value of the proposed approaches for object detection in remote sensing images.

## 1 INTRODUCTION

With the continuous advancement of remote sensing imaging technologies and deep learning algorithms, deep learning–based object detection in remote sensing images has demonstrated significant application potential in fields such as urban governance, smart agriculture, and national defense security[1]. However, in real-world complex scenarios, this technology still faces multiple challenges. First, due to the long imaging distance of satellite and aerial platforms, targets in high-resolution remote sensing images usually occupy only a very small number of pixels (most are smaller than $32 \times 32$ pixels), which makes effective feature extraction extremely difficult. Second, remote sensing images are characterized by complex and diverse background structures, where urban buildings, transportation facilities, farmland, and natural terrain are interwoven. This complexity reduces the contrast between targets and the background and easily leads to feature confusion. Third, remote sensing scenes contain a wide variety of object categories with significant appearance variations. Even within the same category, substantial differences in scale, shape, and texture may exist, further increasing the difficulty of model generalization across diverse scenarios.

In addition to small object detection, multi-scale variation is one of the most fundamental challenges in remote sensing imagery. Objects in aerial images exhibit significant scale differences, ranging from vehicles that occupy only a few pixels to large buildings spanning hundreds of pixels, resulting in a substantial scale span. This wide variation makes it difficult for traditional feature pyramid structures to simultaneously preserve fine-grained spatial details and high-level semantic information. Therefore, enhancing cross-scale feature interaction and adaptive multi-level fusion is crucial for achieving robust multi-scale object detection.

Although detection performance has been continuously improved in recent years through advances in network architectures, feature enhancement strategies, and optimized training mechanisms, issues such as missed detection of small objects and false detections in complex backgrounds remain prominent due to the limited feature representation capability[2]. Therefore, research efforts focusing on enhancing feature representation, multi-scale feature fusion, and improving model robustness have become key directions for further advancing remote sensing object detection technologies[3].

Deep learning–based object detection methods have evolved from early two-stage approaches to lightweight, high-speed, and Transformer-driven architectures[14]. In practical applications, researchers typically select appropriate models according to specific scenario requirements: two-stage detectors are preferred when high accuracy is required; single-stage detectors such as YOLO[22] and SSD[23] are more commonly adopted when real-time performance is prioritized; and in scenarios with complex backgrounds and pronounced long-range dependencies, Transformer-based detectors are increasingly demonstrating greater potential.

Researchers have conducted extensive studies to improve the application of YOLO-based models for object detection in remote sensing images. Xu et [18] enhancedet al. [19] intet al. [20] incorporated Transformer modules into the backbone network to strengthen global dependency modeling and constructed a bidirectional weighted feature pyramid network (FPN) in the neck, enabling adaptive cross-scale fusion of disease-related features. In addition, Zeng et al. [21] proposed a balanced feature pyramid network based on atrous spatial pyramid pooling, which effectively integrates multi-scale feature information and further enhances detection accuracy. Although these studies have made significant progress in addressing challenges such as object diversity and complex backgrounds in remote sensing images, there remains considerable room for improvement in handling weak target features and achieving higher detection accuracy for small-sized objects.

To address the aforementioned issues, this paper proposes two improved lightweight object detection models based on YOLOv11n. By introducing the Large Separable Kernel Attention (LSKA) mechanism to enhance global contextual perception and incorporating the gather-and-distribute structure of Gold-YOLO to achieve more effective multi-scale feature fusion, the proposed methods aim to significantly improve detection accuracy in complex remote sensing scenes while preserving the lightweight and real-time advantages of the original model. Extensive experiments conducted on the DOTA-v1 dataset demonstrate the effectiveness and practical value of the proposed approaches.

## 2. YOLOv11n Model

YOLOv11n[12] is a lightweight single-stage object detector designed to balance detection accuracy and computational efficiency. Compared with earlier YOLOv8[5] versions, it introduces improved feature extraction and lightweight design strategies, making it suitable for real-time applications. However, when applied to high-resolution remote sensing images, YOLOv11n still exhibits limitations in global contextual modeling and cross-scale feature interaction, particularly for small objects and densely distributed targets. These limitations motivate further architectural enhancements tailored to remote sensing scenarios.

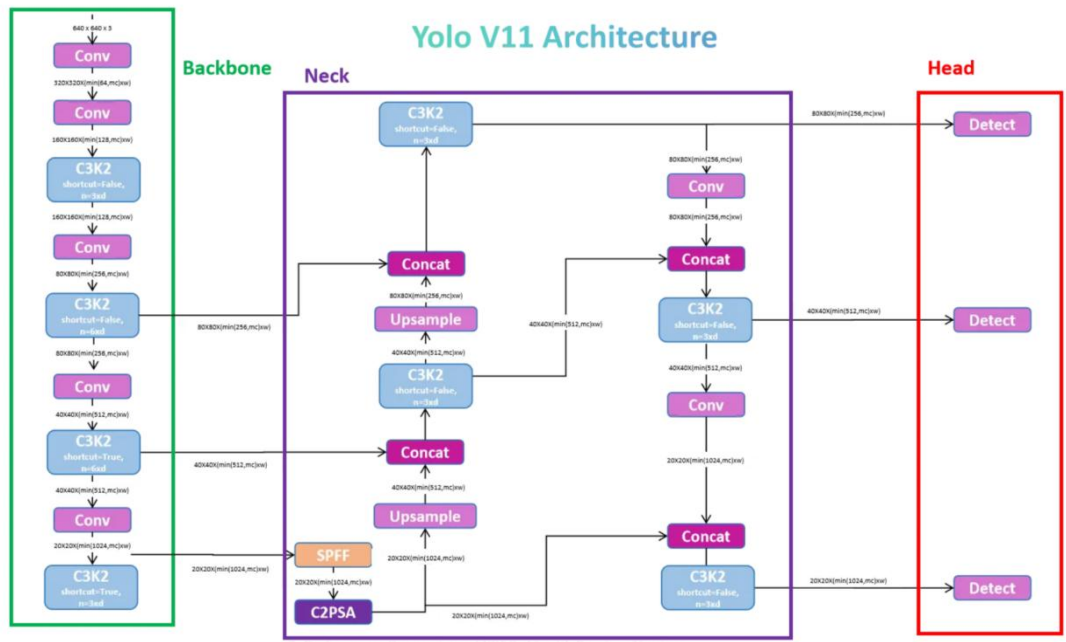

*Figure 1. YOLOv11 Architecture[13]*

## 3. GOLD-YOLO

Gold-YOLO[6] is an efficient object detection framework designed to improve multi-scale feature fusion through a gather-and-distribute (GD) mechanism[7]. By jointly aggregating features from different network depths and redistributing the fused information across layers, Gold-YOLO enhances cross-scale information flow while maintaining low latency and computational efficiency. This design is particularly beneficial for object detection tasks involving large scale variations and dense object distributions, which are common in remote sensing imagery.

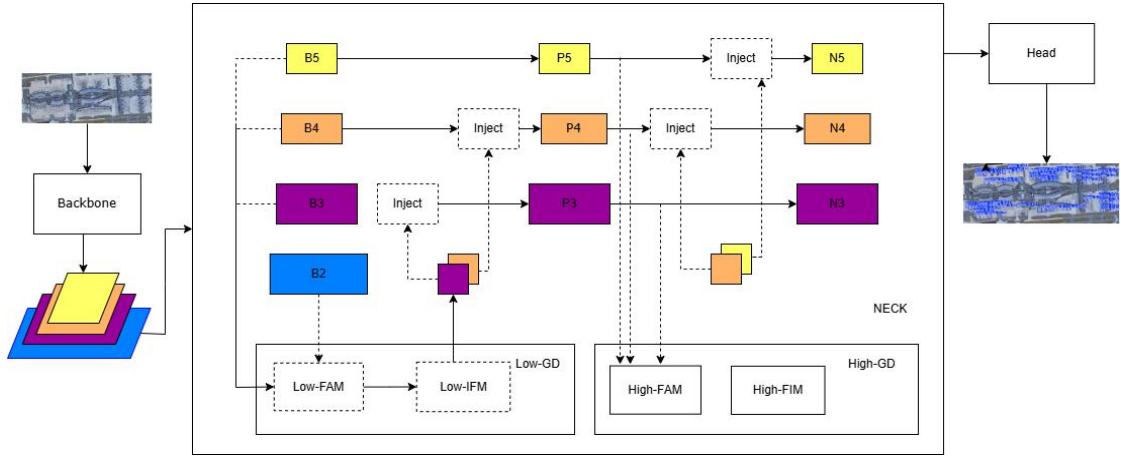

*Figure 2. Gold-YOLO architecture*

As illustrated in Fig. 2, the Gold-YOLO architecture consists of a backbone network, a GD-based neck, and a detection head. The GD structure is composed of two complementary branches: a low-level gather-and-distribute (Low-GD) branch and a high-level gather-and-distribute (High-GD) branch. The Low-GD branch focuses on processing shallow, high-resolution feature maps to preserve fine-grained spatial details that are critical for small object detection, while the High-GD branch emphasizes deep semantic features to enhance contextual understanding for medium and large objects.

Within each branch, feature alignment and information fusion are performed to alleviate feature misalignment caused by scale differences. Specifically, feature alignment modules (FAM) are employed to spatially align features from adjacent layers, followed by information fusion modules (IFM) that integrate aligned features to generate more consistent representations[8]. The fused features are then injected back into different feature levels through the Inject module, enabling effective information redistribution and enhancement across the network.

In addition, Gold-YOLO incorporates a Lightweight Adjacent Fusion (LAF) module to further strengthen feature interaction between neighboring layers[15]. By performing scale alignment via pooling and up/down-sampling and adopting lightweight fusion operations, LAF allows each feature level to receive complementary information from adjacent layers with minimal computational overhead. Through the coordinated operation of the GD mechanism and LAF module, Gold-YOLO achieves more robust and consistent multi-scale feature representations, providing a solid foundation for improving detection performance in complex remote sensing scenes.

## 4. Large Separable Kernel Attention Mechanism

Large Separable Kernel Attention (LSKA) is an efficient attention mechanism designed to capture large receptive fields while maintaining low computational complexity[16]. In remote sensing images, objects are often surrounded by complex backgrounds and exhibit weak visual contrast, making global contextual information particularly important for accurate detection. However, directly employing large-kernel convolutions leads to a significant increase in parameters and computational cost, which is unsuitable for lightweight detection models.

LSKA addresses this issue by decomposing large two-dimensional convolution kernels into a series of separable one-dimensional convolutions along the horizontal and vertical directions, combined with depthwise and dilated convolution operations. This design enables the network to approximate large receptive fields efficiently while significantly reducing parameter count and computational overhead. Compared with conventional large-kernel attention mechanisms, LSKA achieves comparable global feature modeling capability with improved efficiency and stability[9].

By generating spatial attention maps based on large-context perception and reweighting the original feature maps, LSKA enhances the network's ability to focus on salient regions and suppress background interference. Owing to its favorable balance between global contextual modeling and lightweight design, LSKA is well suited for integration into YOLOv11n to improve feature extraction performance for small objects and dense scenes in remote sensing imagery.

## 5. MultiSEAMHead

Although the original YOLOv11n detection head achieves high inference efficiency, its ability to jointly exploit multi-level semantic and fine-grained features remains limited. Shallow features often lack sufficient semantic information, while deep features tend to lose spatial details, which negatively affects detection performance in complex remote sensing scenes characterized by dense object distributions, occlusion, and background clutter.

To address this limitation, the MultiSEAMHead(Fig. 3) module enhances the detection head by incorporating multi-level feature fusion and attention-based feature modulation[17]. It leverages depthwise separable convolutions and cross-layer connections to improve information interaction across different feature scales, enabling the detector to better utilize both detailed spatial information and high-level semantic representations.

Furthermore, MultiSEAMHead introduces channel and spatial mixing mechanisms to model inter-channel dependencies and emphasize informative regions. By adaptively reweighting features through attention-based regulation, the detection head becomes more robust to scale variation and background interference. As a result, MultiSEAMHead effectively improves detection accuracy for small and occluded objects while preserving the lightweight characteristics and real-time performance of the YOLOv11n framework.

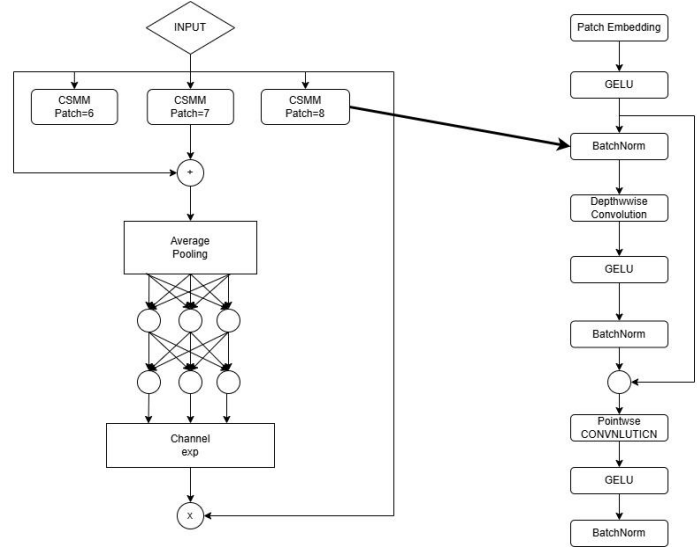

*Figure 3. MultiSEAMHead Structure Diagram*

## 6. Model design

Although YOLOv11n introduces self-attention and lightweight convolutional designs, its architecture is primarily optimized for general-purpose object detection tasks. When applied to remote sensing images, several inherent limitations become apparent. First, the backbone network lacks sufficient large-receptive-field modeling capability, which restricts its ability to capture global contextual information crucial for distinguishing small objects from complex backgrounds. Second, the original neck structure provides limited cross-layer information interaction, leading to suboptimal feature fusion for objects with large scale variations. Finally, the detection head focuses mainly on efficiency, while the joint utilization of multi-level semantic and fine-grained features remains insufficient for robust detection in dense and cluttered scenes.

Motivated by these observations, we introduce targeted architectural modifications to YOLOv11n. The LSKA mechanism is employed to enhance global feature perception with minimal computational overhead, addressing the limited receptive field issue. The Gold-YOLO neck is integrated to strengthen multi-scale feature aggregation and distribution, improving information flow across different feature levels. Furthermore, the MultiSEAMHead is adopted to enhance the detection head's ability to jointly model spatial and channel-wise dependencies. These modifications are specifically designed to overcome the challenges of remote sensing object detection while preserving the lightweight nature of YOLOv11n.

### 6.1 YOLOv11n-LSKA-GoldYOLO

First improved lightweight object detection model proposed in this study, YOLOv11n-LSKA-GoldYOLO, is built upon the YOLOv11n framework (Fig. 4). To enhance feature representation in complex scenes, a Large Kernel Separable Attention (LSKA) mechanism is introduced into the backbone network. LSKA employs a decomposed large convolution kernel structure that effectively enlarges the receptive field while keeping computational costs manageable, enabling the model to capture global contextual information more comprehensively and suppress redundant features. This design provides significant advantages for small object recognition and dense scene detection.

During the feature fusion stage, the Gold-YOLO Neck structure is further employed to achieve more efficient multi-scale information interaction and adaptive feature integration, thereby improving feature propagation and detection accuracy across objects of different scales. By combining the strengths of the LSKA mechanism and the Gold-YOLO Neck, YOLOv11n-LSKA-GoldYOLO significantly enhances detection performance while maintaining low computational cost and real-time inference capability, offering a practical and efficient solution for lightweight object detection tasks.

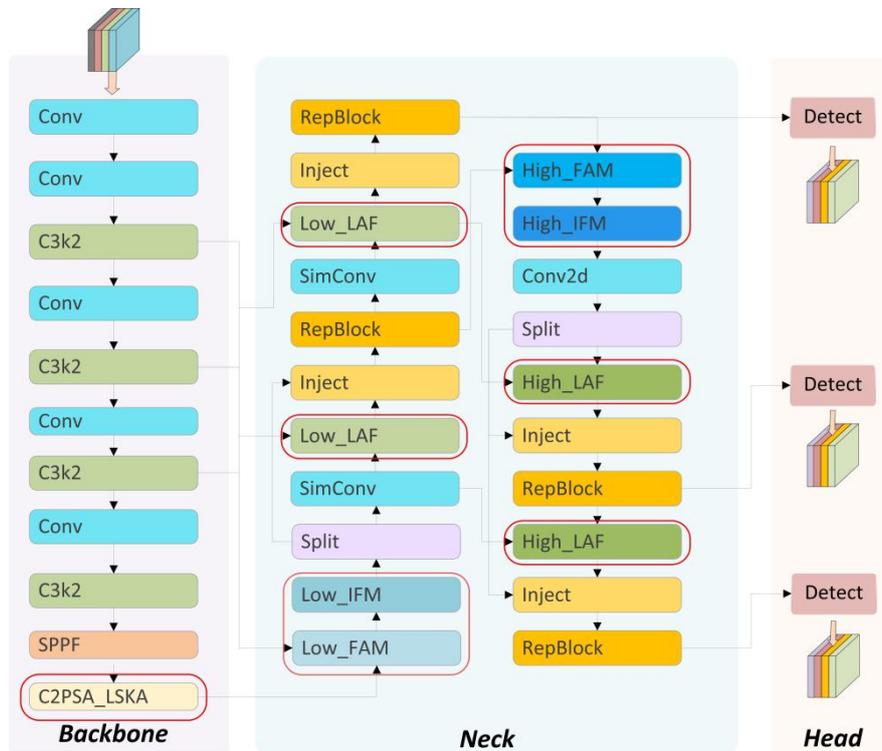

*Figure 4. YOLOv11n-LSKA-GoldYOLO*

### 6.2 YOLOv11n-GoldYOLO-MultiSEAMHead

This study further proposes the lightweight object detection model YOLOv11n-GoldYOLO-MultiSEAMHead, whose overall architecture is based on YOLOv11n and incorporates the efficient feature fusion design of Gold-YOLO (Fig. 5). During the feature interaction stage, the model employs the Gold-YOLO Neck structure to facilitate more comprehensive multi-scale information flow and a more stable adaptive fusion mechanism, enabling efficient feature propagation between shallow and deep layers and improving detection performance across objects of varying scales.

To further enhance the representational capacity of the detection head, the model integrates the MultiSEAMHead module. This module performs joint modeling of multi-level features through depthwise separable convolutions, multi-scale feature extraction, and cross-layer connections, allowing the network to more comprehensively capture fine-grained details and high-level semantic information. By leveraging its built-in Channel and Spatial Mixing Units and globally context-aware attention weighting, MultiSEAMHead enables the network to accurately focus on key regions and suppress irrelevant features in scenarios involving occlusion, complex backgrounds, or dense objects. Additionally, the improved regression loss in the detection head further stabilizes the training process and accelerates the convergence of bounding box predictions.

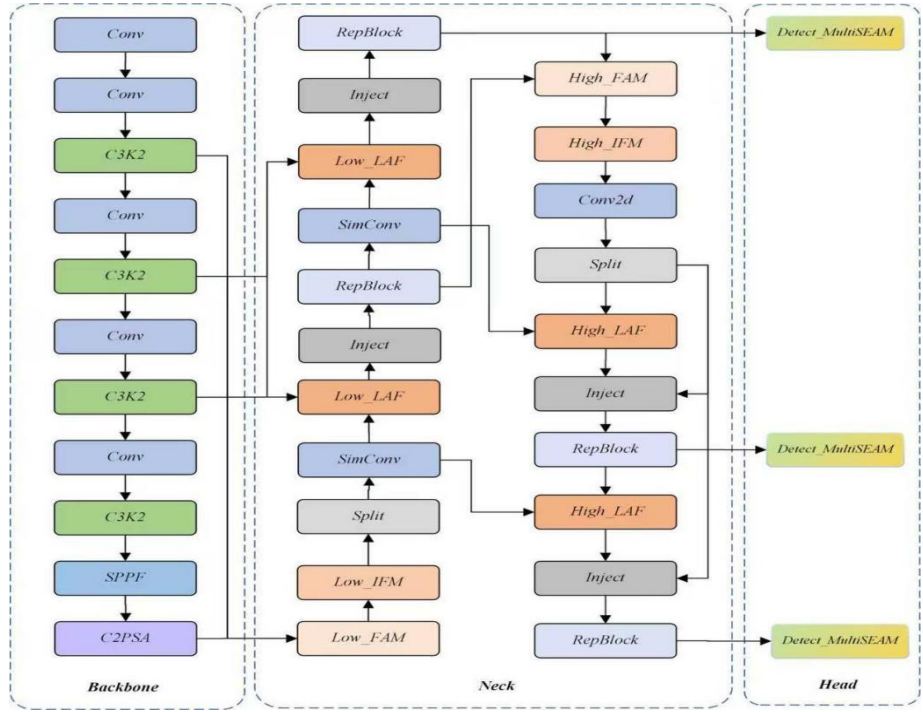

*Figure 5. YOLOv11n-GoldYOLO-MultiSEAMHead*

## 7. Experimental Procedure
### 7.1 Experimental Dataset

The experiments in this study use the publicly available DOTA-v1 (Dataset for Object deTection in Aerial Images Version 1) dataset[11], which is one of the most representative benchmarks for aerial image object detection in the remote sensing domain. DOTA-v1 is specifically designed for multi-class object detection tasks in high-resolution remote sensing images. Released by the Aerospace Information Research Institute of the Chinese Academy of Sciences, the dataset includes images from various aerial and satellite imaging platforms, such as Google Earth, GF-2, and JL-1, offering wide coverage, diverse data types, and complex geometric structures.

The dataset contains 2,806 high-resolution images, with resolutions ranging from 800×800 to 4,000×4,000 pixels and spatial resolutions covering 0.1–1 m/px. It encompasses both dense urban areas with buildings and transportation infrastructure, as well as natural regions including farmland, ports, rivers, and forests, providing high scene diversity and complexity. The dataset includes 15 typical object categories (Fig. 6).

To increase task difficulty and ensure comprehensive algorithm evaluation, the object scale distribution in DOTA-v1 is highly imbalanced. It contains numerous small objects of only a few dozen pixels as well as large objects spanning several hundred pixels. Many areas also feature severe occlusion, dense object distributions, and background interference, making DOTA-v1 an important benchmark for assessing model robustness and generalization ability.

For the experimental setup, DOTA-v1 is divided into training (train), validation (val), and testing (test) sets with an 8:1:1 split. The training set is used for model learning, the validation set for parameter tuning and early stopping, and the test set for final performance evaluation.

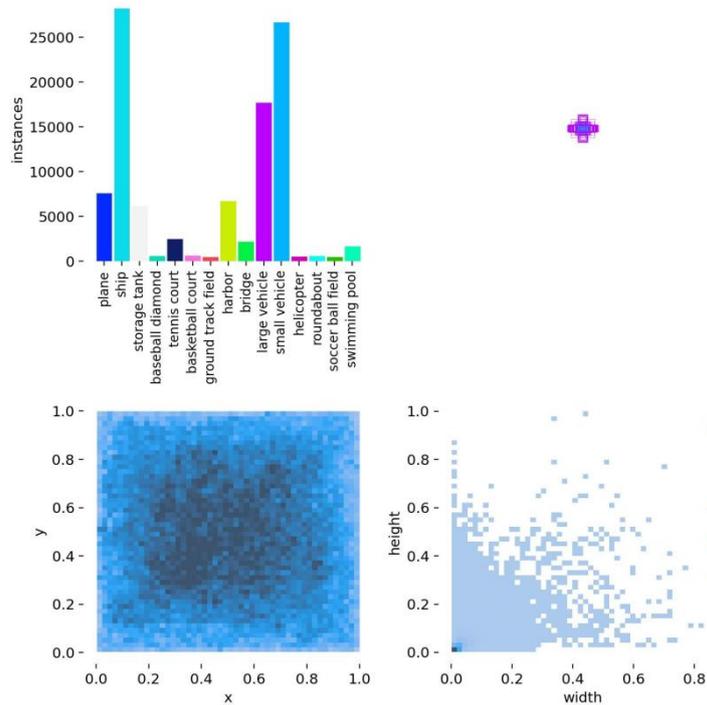

*Figure 6. Experimental Dataset*

**7.2 Experimental Environment**

The experiments in this study were conducted on the AutoDL platform, leveraging high-performance hardware and a stable software environment. The specific configuration is as follows:

- Hardware: NVIDIA RTX 4090 GPU (24 GB) to accelerate model training, paired with an Intel Core i9-14900KF processor and 60 GB of RAM to ensure efficient and stable training.

- Software: PyTorch 2.1.0 with CUDA 12.1 support as the primary deep learning framework, using Python 3.10 for programming.

The experimental setup includes 300 training epochs, with the batch size automatically adjusted according to GPU memory capacity (batch = -1). The Stochastic Gradient Descent (SGD) optimizer was used, while all other training parameters were kept at their default settings.

**7.3 Evaluation Metrics**

Precision measures the proportion of correctly predicted positive samples among all samples predicted as positive, reflecting the reliability of the model's positive predictions. It is calculated as follows:

$$P = \frac{TP}{TP + FP}$$

Recall measures the model's ability to capture all true positive samples in the dataset, i.e., how many of the actual positive instances are correctly identified by the model. It is calculated as:

$$R = \frac{TP}{TP + FN}$$

Mean Average Precision (mAP) is used to comprehensively evaluate the detection accuracy of an object detection model across all classes. It is defined as the mean of the Average Precision (AP) values for each class, generally computed as:

$$mAP = \frac{1}{N}\sum_{i=1}^{N} AP_i$$

Here, TP, FP, and FN respectively denote the numbers of correctly detected bounding boxes, false positive boxes, and missed boxes. AP represents the area enclosed by the P-R (Precision-Recall) curve and the coordinate axes. mAP50 (i.e., mean Average Precision at IoU = 0.50) is calculated based on an IoU threshold of 0.50 and is used to measure the model's detection accuracy at this threshold; whereas mAP50-95 is computed over IoU values from 0.50 to 0.95 (with a step size of 0.05), providing a more comprehensive reflection of the model's detection performance under different strictness levels and offering greater reference value for evaluating the model's generalization ability and robustness.

**8. Experimental Results**

In this study, based on YOLOv11n, two improved models—YOLOv11n-LSKA-GoldYOLO and YOLOv11n-GoldYOLO-MultiSEAMHead—were constructed by introducing the Large Separable Kernel Attention (LSKA) mechanism, the Gold-YOLO Neck structure, and the MultiSEAMHead detection head. Comparative experiments were conducted against the original YOLOv11n model, and ablation studies were performed to evaluate the contribution of each improved module. The performance of the new models on the DOTA-v1 dataset was thoroughly assessed.

| Модель | Recall% | mAP50% | mAP50-95% |
|---|---|---|---|
| YOLOv8n | 36.6 | 39.9 | 24.2 |
| YOLOv9t | 37 | 41.3 | 25.5 |
| YOLOv10n | 35.8 | 38.6 | 23.4 |
| YOLOv11n | 40.2 | 42 | 25.7 |
| YOLOv11n+LSKA | 39 | 42.9 | 26 |
| YOLOv11n+goldyolo | 40.6 | 42.2 | 25.8 |
| YOLOv11n+MultiSEAMHead | 37.4 | 40.4 | 24.2 |
| YOLOv11n+MultiSEAMHead+goldyolo | 40.3 | 43.8 | 26.3 |
| YOLOv11n+LSKA+goldyolo | 38.8 | 43.3 | 26 |

*Table 1.Comparison results*

Table 1 summarizes the key performance metrics, including mAP50%, mAP50-95%, and Recall. The experimental results show that the YOLOv11n-LSKA-GoldYOLO model achieves a 1.3% improvement in mAP50% and a 0.3% improvement in mAP50-95% over YOLOv11n. The YOLOv11n-GoldYOLO-MultiSEAMHead model achieves a 1.8% improvement in mAP50% and a 0.6% improvement in mAP50-95%. The synergistic effect of the different modules not only enhances the diversity of feature representations but also strengthens overall model robustness, enabling higher stability and accuracy in complex scene detection tasks.

Figure 7 shows original images from the DOTA-v1 dataset and detection results of the YOLOv11n model. Figure 8a presents the detection results of the YOLOv11n-LSKA-GoldYOLO model on the validation set, while Figure 8b shows the detection results of the YOLOv11n-GoldYOLO-MultiSEAMHead model on the validation set.

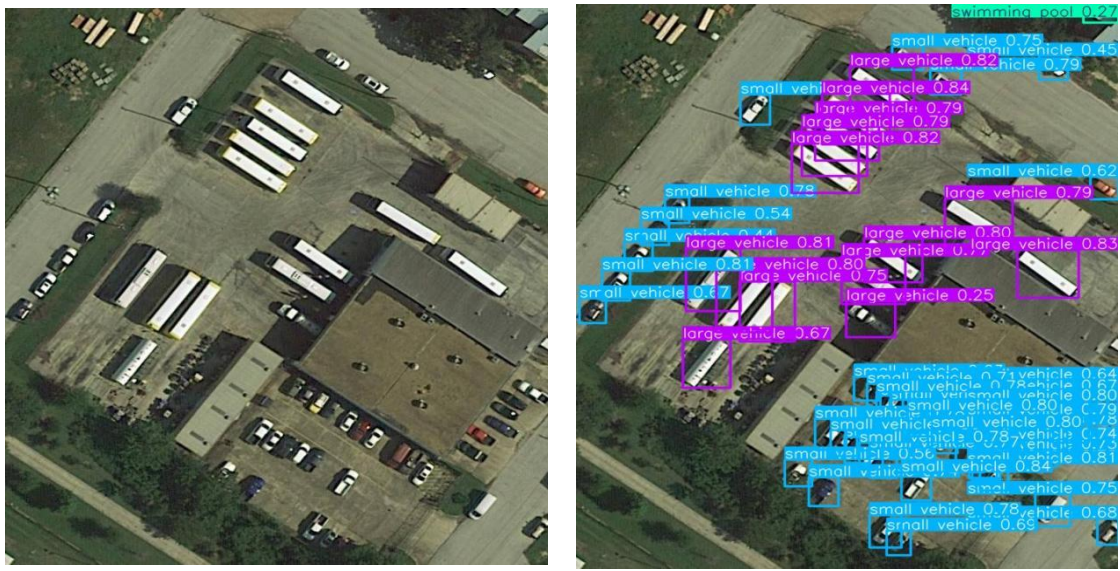

*7a*  *7b*

*Figure 7 Original image and YOLOv11n Detection Results*

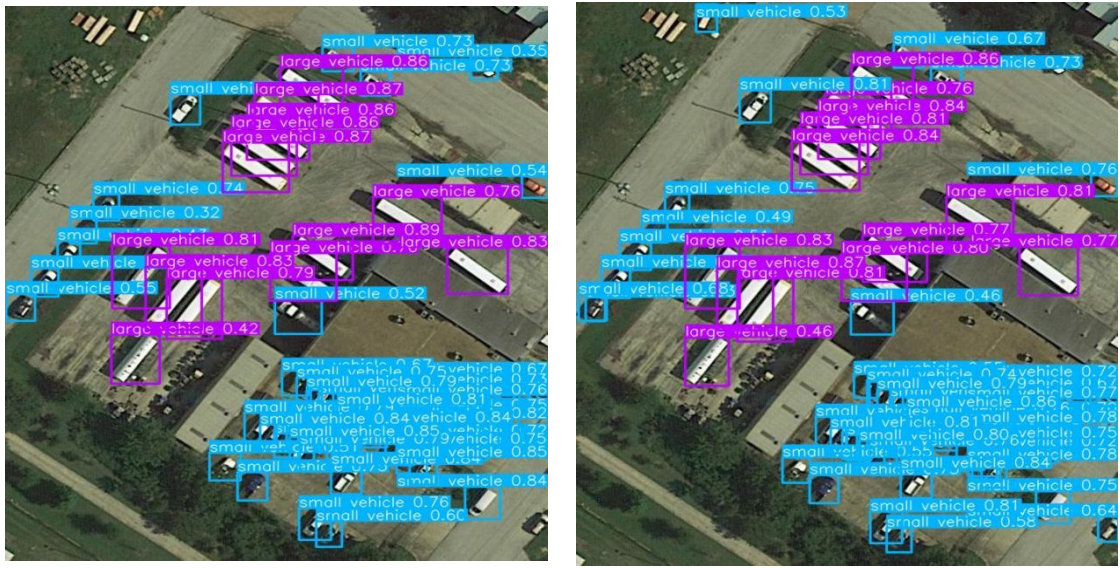

*Figure 8. YOLOv11n-LSKA-GoldYOLO Detection Results(a) and YOLOv11n-GoldYOLO-MultiSEAMHead Detection Results(b)*

Both proposed models outperform the baseline YOLOv11n in detection performance, yet they emphasize different architectural designs and practical advantages. YOLOv11n-LSKA-GoldYOLO enhances the backbone network by introducing a large-receptive-field attention mechanism, making it particularly suitable for tasks where global contextual information is critical, such as scenarios involving small, sparsely distributed targets or objects embedded in complex backgrounds. Since it introduces relatively fewer modifications to the detection head, it incurs lower computational overhead and is therefore more appropriate for deployment environments with strict real-time requirements or limited computational resources. In contrast, YOLOv11n-GoldYOLO-MultiSEAMHead focuses more on strengthening feature interaction during the detection stage. By enhancing multi-level semantic fusion and spatial–channel attention modulation, this model demonstrates greater robustness in scenarios with dense targets, occlusion, or significant scale variations. Therefore, it is better suited for high-precision detection tasks in urban or port environments where object overlap and background clutter are common. By providing two targeted variants, we enable flexible selection of a more appropriate model according to specific application requirements.

## 9. Conclusion

This study addresses the challenges of large-scale variation, dense distribution, and complex backgrounds in aerial remote sensing image object detection by proposing improved lightweight object detection models based on YOLOv11n, namely YOLOv11n-LSKA-GoldYOLO and YOLOv11n-GoldYOLO-MultiSEAMHead. By introducing the Large Separable Kernel Attention (LSKA) mechanism into the backbone network, the models significantly enhance global receptive field and contextual information modeling capabilities. Incorporating the Gold-YOLO Neck during feature fusion enables more efficient and stable multi-scale feature interactions, and the integration of the MultiSEAMHead module achieves deep cross-layer feature fusion with stronger spatial–channel joint modeling capability.

Experimental results on the DOTA-v1 dataset demonstrate that the proposed improved models outperform the baseline YOLOv11n across key metrics, confirming the effectiveness of each module in enhancing feature representation and detection accuracy. In practical applications, YOLOv11n-LSKA-GoldYOLO is more suitable for lightweight deployment scenarios requiring enhanced global contextual perception, whereas YOLOv11n-GoldYOLO-MultiSEAMHead is preferable for complex and densely distributed scenes demanding higher detection robustness. The two models provide flexible and extensible solutions for different remote sensing application requirements.

## CONFLICT OF INTEREST

The authors declare that they have no conflicts of interest.

## DATA AVAILABILITY STATEMENTS

The data underlying this article will be shared on reasonable request to the corresponding author.

## DECLARATION OF GENERATIVE AI

During the preparation of this work, the authors used ChatGPT with the aim of improving the language. Following its use, the authors reviewed and edited the content as needed and take full responsibility for the content of the publication.